\documentclass[runningheads]{llncs}
\usepackage{dsfont}
\usepackage{mathtools}
\usepackage{graphicx}
\usepackage{stmaryrd}
\usepackage{bm}
\usepackage{appendix}
\usepackage{comment}
\usepackage{url}

\makeatletter
\def\blfootnote{\xdef\@thefnmark{}\@footnotetext}
\makeatother

\def\proof{\noindent\mbox{\sc proof: }}
\def\eproof{\rm\hspace*{\fill}$\rule{7pt}{7pt}$\vspace{10pt}}

\def\C#1{\text{CV}_{#1}}
\def\C#1{\text{CV}_{#1}}
\def\Pthree{P\textsuperscript{3}\,}
\def\tTPR{\widehat{\text{TPR}}}
\def\tFPR{\widehat{\text{FPR}}}
\def\TPR{{\text{TPR}}}
\def\CTpr{\text{CV}_{\text{TPR}}}
\def\CFpr{\text{CV}_{\text{FPR}}}
\def\SigmaTpr{{\sigma_{\text{TPR}}}}
\def\SigmaFpr{{\sigma_{\text{FPR}}}}

\def\FPR{{\text{FPR}}}
\def\Prec{{\text{Prec}}}
\def\Prob{{\text{Prob}}}
\def\tPrec{\widehat{\text{Prec}}}

\def\ITpr{\cal{I}_{\text{TPR}}}
\def\IFpr{\cal{I}_{\text{FPR}}}
\def\IPrec{{\cal{I}_{\text{Prec}}}}
\def\LB{{\text{LB}}}
\def\UB{{\text{UB}}}

\begin{document}
\title{On Model Evaluation under Non-constant Class Imbalance}
\author{Jan Brabec\inst{1,2} \and
Tom\'{a}\v{s} Kom\'{a}rek\inst{1,2} \and
Vojt\v{e}ch Franc\inst{2} \thanks{VF was supported by OP VVV project CZ$.02.1.01\backslash 0.0\backslash 0.0\backslash 16\_019\backslash 0000765$ Research Center for Informatics.} \and
Luk\'{a}\v{s} Machlica\inst{3}}
\authorrunning{J. Brabec et al.}
\institute{Cisco Systems, Inc., Karlovo Namesti 10 Street, Prague, Czech Republic \\ \email{\{janbrabe,tomkomar\}@cisco.com} \and
Czech Technical University in Prague, Faculty of Electrical Engineering, Czech Rep. \\ \email{xfrancv@cmp.felk.cvut.cz}
\and
Resistant.ai, Prague, Czech Republic \\
\email{lukas.machlica@resistant.ai}}
\maketitle              % typeset the header of the contribution
%%

%Many real-world classification problems are significantly class-imbalanced to detriment of the class of interest. The standard set of proper evaluation metrics is well-known but it is often applied blindly without focus on the working point at which the system should operate once deployed. This is essential for problems solved in applied machine learning, because the imbalance ratios may greatly affect the interpretation and understanding of the results if not handled properly, and therefore may negatively impact assessment of proposed methods. Classifiers with seemingly high efficacy on a particular test dataset may completely fail once deployed into the wild for reasons that would become immediately apparent if the evaluation on the test set was performed with the real application in mind, not in isolation. 
%In this paper, we argue that the main focus in applied machine learning should be given to the evaluation metric itself. We review two core evaluation metrics, ROC and PR curves, point out the caveats that may lead to wrong conclusions, and propose several alternatives providing different views beyond a particular test dataset that may reveal efficacy issues before the system is deployed.
\begin{abstract}
Many real-world classification problems are significantly class-imbalanced to detriment of the class of interest. The standard set of proper evaluation metrics is well-known but the usual assumption is that the test dataset imbalance equals the real-world imbalance. In practice, this assumption is often broken for various reasons. The reported results are then often too optimistic and may lead to wrong conclusions about industrial impact and suitability of proposed techniques. We introduce methods\footnote{Supplementary code related to techniques described in this paper is available at: \url{https://github.com/CiscoCTA/nci_eval}} focusing on evaluation under non-constant class imbalance. We show that not only the absolute values of commonly used metrics, but even the order of classifiers in relation to the evaluation metric used is affected by the change of the imbalance rate. Finally, we demonstrate that using subsampling in order to get a test dataset with class imbalance equal to the one observed in the wild is not necessary, and eventually can lead to significant errors in classifier's performance estimate. 

\end{abstract}

%\blfootnote{VF was supported by OP VVV project CZ$.02.1.01\backslash 0.0\backslash 0.0\backslash 16\_019\backslash 0000765$ Research Center for Informatics.}
%
%
%

% \blfootnote{Code available at: https://github.com/anonymized-path-for-review-purposes}

\section{Introduction}

Class-imbalanced problems arise if number of samples in one of the classes, often in the class of interest, is significantly lower than in the other class, often the background class. Such problems are present in variety of different domains such as medicine \cite{rahman2013addressing}, finance \cite{Phua:2004:MRF:1007730.1007738,wei2013effective,yu2010multiscale}, cybersecurity \cite{axelsson2006base,bayesianforests,damodaran2017comparison} and many others.

In highly imbalanced problems it is essential to use suitable evaluation metrics to correctly assess the merit of pursued algorithms and realistically judge their impact before they are deployed into the wild. Methods for evaluation of classifiers on class-imbalanced datasets are well known and have been thoroughly described in the past \cite{chawla2009data,he2009learning,kotsiantis2006handling,sokolova2009systematic}.

It is usually assumed that the imbalance of the test dataset is the same as in the real distribution on which the model will operate once deployed into production environment. However, this assumption is often broken, because of different reasons ranging from selection bias when constructing the test dataset, high costs of acquiring large dataset mainly in situations when the imbalance is high (e.g. $1 : 10^4$), to the fact that often not a single general distribution exists (e.g. disease classifier may face different priors depending on the location).%, \ldots).
%The class imbalance directly affects the precision metric, and this discrepancy between imbalances in test dataset and real world is often the root cause of too optimistic results and wrong interpretation of the impact in industrial applications.

Discrepancy between imbalances in test datasets and real world is often the root cause of too optimistic results leading to wrong expectations of the impact in industrial applications. This is detrimental to the research community, because it creates confusion about which problems are still open and which are solved. It might discourage groups from working on such problems, and make it harder for researchers still investigating the field to convince the community that in the light of the too optimistic prior work their results have still impact.

Throughout this paper, we frame and investigate the problem of classifier evaluation dropping the assumption of constant class imbalance.% and introduce methods suited for this task.
We focus on precision related metrics as one of the most popular metrics for imbalanced problems \cite{chawla2009data,he2009learning}. We show how these metrics can be computed for arbitrary class imbalances and any test dataset without the need to re-sample the data. We also inspect their behavior as a function of the imbalance rate. We show that Precision-Recall (PR) curves have little value without stating the corresponding imbalance ratio which can dramatically affect the results and their assessment.

We demonstrate that change in imbalance rate, maybe surprisingly, affects also the ranking of classifiers under these metrics. We argue that instead of tabulating the results for a single dataset, it is beneficial to plot the dependence on the class imbalance rate whenever possible. Such plots provide considerably more information for wider audience.

We also describe how errors in measurements can be assessed and that they can significantly affect the reliability of measured precision mainly in cases when low regions of false positive rate are of interest. This can be primarily attributed to the fact that the test dataset is finite. Therefore, we further elaborate how the class imbalance increases the demands on the size of test dataset.

Most importantly, {\it we refute the common understanding that the best practice is to alter the test dataset so that class imbalance matches the imbalance of the pursued distribution} as is suggested e.g. in \cite{pendlebury2019tesseract}. We show how re-sampling of a dataset may lead to significant errors in measurements. We stress that the test dataset should be constructed in a way to allow measurements of false-positive and true-positive rates with  errors as small as possible. We show that the crucial entity to focus on is the coefficient of variation related to both true-positive and false-positive rates.

\section{Preliminaries}
\label{sec:preliminaries}

Throughout this paper we are concerned with the binary classification task. Let $\bm{x} \in \mathcal{X}$ be an input and $y \in \mathcal{Y} = \left\{-1, 1\right\}$ be a target. We call the class $y=-1$ negative class and the class $y=1$ positive class. The positive class is assumed to be the minority class and the negative class is the majority class. We do not assume that there exists a single real-world joint-probability distribution $p(\bm{x}, y)$ but instead consider a parametric family:

\begin{equation}
p(\bm{x}, y; \eta) = p(\bm{x} | y) \cdot P(y; \eta)\text{, where }P(y; \eta) = \begin{cases}
1 - \eta & y = -1\\
\eta & y = 1
\end{cases}.
\end{equation}

Parameter $\eta \in \left[0, 1\right]$ specifies the positive class prevalence. If we consider a classifier $h: \mathcal{X} \mapsto \mathcal{Y}$ then the following classifier evaluation metrics can be expressed as probabilities:

\begin{equation}
\label{eq:tpr}
    \text{TPR} = \text{Recall} = P(h(\bm{x}) = 1 | y = 1) = \mathds{E}_{\bm{x}\sim p(\bm{x}|y=1)} [ h(\bm{x}) = 1 ]
\end{equation}
\begin{equation}
\label{eq:fpr}
    \text{FPR} = P(h(\bm{x}) = 1 | y = -1) = \mathds{E}_{\bm{x}\sim p(\bm{x}|y=-1)} [ h(\bm{x}) = 1 ]
\end{equation}
\begin{equation}
\label{eq:prec}
    \text{Prec}(\eta) = P(y = 1 | h(\bm{x}) = 1) = \frac{\text{TPR} \cdot \eta}{\text{TPR} \cdot \eta + \text{FPR} \cdot (1 - \eta)}
\end{equation}

TPR stands for true-positive-rate (also called recall or sensitivity), FPR for false-positive-rate and Prec for precision. Formula \eqref{eq:prec} is derived using Bayes’ theorem. We can observe that both TPR and FPR are not affected by the positive class prevalence but precision is. This observation is very important for the rest of this paper.

To estimate the above-mentioned metrics we need to evaluate the classifier on a test dataset. We assume that the test dataset is sampled i.i.d. from $p(\bm{x}, y; \eta_{test})$ where $\eta_{test}$ may or may not correspond to a positive class prevalence connected to some real-world application of the classifier. $TP$, $FP$, $TN$, $FN$ denote the number of true positives, false positives, true negatives and false negatives, respectively and $N=TP + FP + TN + FN$ equals the size of the test set.

Prevalence of the positive class in the test dataset $p_{+}$ and imbalance ratio (IR) are defined as (one can be computed from the other easily):

\begin{equation}
    p_{+} = \frac{TP+FN}{N}, \:\: IR = \frac{TP+FN}{TN+FP}.
\end{equation}

$\tTPR$ is defined as the fraction of positive samples that were classified correctly:

\begin{equation}
\label{equ:estimatedTPR}
    \tTPR = \frac{1}{|\mathcal{X}^+|} \sum_{\bm{x}\in\mathcal{X}^+} \llbracket h(\bm{x})=1 \rrbracket = \frac{\text{TP}}{|\mathcal{X}^+|} = \frac{\text{TP}}{\text{TP} + \text{FN}}, 
\end{equation}

where $\llbracket \cdot \rrbracket$ is the indicator function. $\tFPR$ is defined as the fraction of negatives samples that were classified incorrectly:

\begin{equation}
    \tFPR = \frac{1}{|\mathcal{X}^-|} \sum_{\bm{x}\in\mathcal{X}^-} \llbracket h(\bm{x})=1 \rrbracket = \frac{\text{FP}}{|\mathcal{X}^-|} = \frac{\text{FP}}{\text{FP} + \text{TN}}.
\end{equation}

$\tPrec$ is the number of true positives out of all the positive predictions:

\begin{equation}
\label{eq:test_prec}
    \tPrec(\eta) = \frac{\tTPR \cdot \eta}{\tTPR \cdot \eta + \tFPR \cdot (1 - \eta)}
\end{equation}

It can be easily shown that $\tPrec(p_{+}) = \text{TP}/(\text{TP} + \text{FP})$ resolves to the standard formula used to compute precision.
It holds that the metrics measured on the test dataset approach their true values originating from the distribution $p(\bm{x}, y; \eta)$ as the size of the dataset grows. In other words $p_{+} \rightarrow \eta_{test}, \tTPR \rightarrow \text{TPR}$, $\tFPR \rightarrow \text{FPR}$ and $\tPrec \rightarrow \text{Prec}$ as $N$ approaches infinity, but the errors in estimation caused by limited size of test dataset are often significant enough to deserve consideration, particularly during classifier evaluation in settings that are heavily class-imbalanced. We elaborate on this in Section \ref{sec:errors}.

% \begin{equation}
%     z = f(x).
% \end{equation}

% \begin{equation}
%     p(z) = \phi_+ p(z|y=1) + \phi_- p(z|y=0).
% \end{equation}

% \begin{equation}
%     \mathrm{TPR}(t) = P[z>t|y=1] = \int_{t}^{\infty} p(z|y=1)\mathrm{d}z.
% \end{equation}

% \begin{equation}
%     \mathrm{FPR}(t) = P[z>t|y=0] = \int_{t}^{\infty} p(z|y=0)\mathrm{d}z.
% \end{equation}

% \begin{equation}
%     \widehat{\mathrm{TPR}}(t) = \frac{1}{|\mathcal{X}^+|} \sum_{x \in \mathcal{X}^+} \mathds{1} \left [ f(x)>t \right] = \frac{\widehat{\mathrm{TP}}(t)}{|\mathcal{X}^+|}.
% \end{equation}

% \begin{equation}
%     \widehat{\mathrm{FPR}}(t) = \frac{1}{|\mathcal{X}^-|} \sum_{x \in \mathcal{X}^-} \mathds{1} \left [ f(x)>t \right] = \frac{\widehat{\mathrm{FP}}(t)}{|\mathcal{X}^-|}.
% \end{equation}

% \begin{itemize}
%     \item independent on class-imbalance
%     \item auc - problem with region of interest / log-scale / specific algorithm
% \end{itemize}

% \begin{equation}
% \begin{aligned}
%     \mathrm{Precision}(t,\phi_+) &= P[y=1|z>t] = \frac{P[z>t|y=1] \phi_+}{P[z>t|y=1] \phi_+ + P[z>t|y=0] \phi_-}\\ &= \dfrac{\phi_+ \mathrm{TPR}(t)}{\phi_+ \mathrm{TPR}(t) + \phi_- \mathrm{FPR}(t)} = \dfrac{\mathrm{TP}(t)}{\mathrm{TP}(t) + \mathrm{FP}(t)}
% \end{aligned}
% \end{equation}

\section{Precision in the light of different class imbalance ratios}
\label{sec:adjust}

Equation \eqref{eq:test_prec} in Section \ref{sec:preliminaries} shows that the class imbalance ratio of the test dataset directly impacts the measured precision. As such, the test dataset class imbalance must be considered when interpreting the results to assess viability of the classifier for a given application.

Fortunately, it is not necessary for a test dataset's imbalance ratio to be equivalent to the real-world imbalance. Equation \eqref{eq:test_prec} shows how to estimate precision ($\tPrec$), that corresponds to any class imbalance, from $\tTPR$ and $\tFPR$ which are estimated from the test dataset and are unaffected by it's imbalance.

In Section \ref{sec:errors} we provide rationale and show that matching the real-world class imbalance is often sub-optimal and not desirable for correct evaluation.

%As an alternative to using Equation \eqref{eq:test_prec}, precision for any imbalance rate can be computed by the following formula that uses the positive class prevalence of the test dataset ($p_{+}$) directly and does not use $\tTPR$ and $\tFPR$. As such it is a natural extension of the well-known precision formula that uses TP and FP directly:

%\begin{equation}
%    \label{eq:adjPrec1}
%    \tPrec(\eta) = \frac{\frac{\eta}{p_{+}} \cdot \text{TP}}
%    {\frac{\eta}{p_{+}} \cdot \text{TP} + \frac{1 - \eta}{1 - p_{+}} \cdot \text{FP}},
%\end{equation}

%This formula is equivalent to \eqref{eq:test_prec}, but offers additional insight into how the precision computed on test dataset may change with the change of class imbalance and how it depends on the prevalence of the positive class $p_{+}$. Because of the equivalence, we will stick to the use of \eqref{eq:test_prec} in the rest of the paper.

\subsection{Positive-prevalence precision curve}
\label{sec:p3}

\begin{figure}[t!]
    \centering
    {\includegraphics[width=0.9\linewidth]{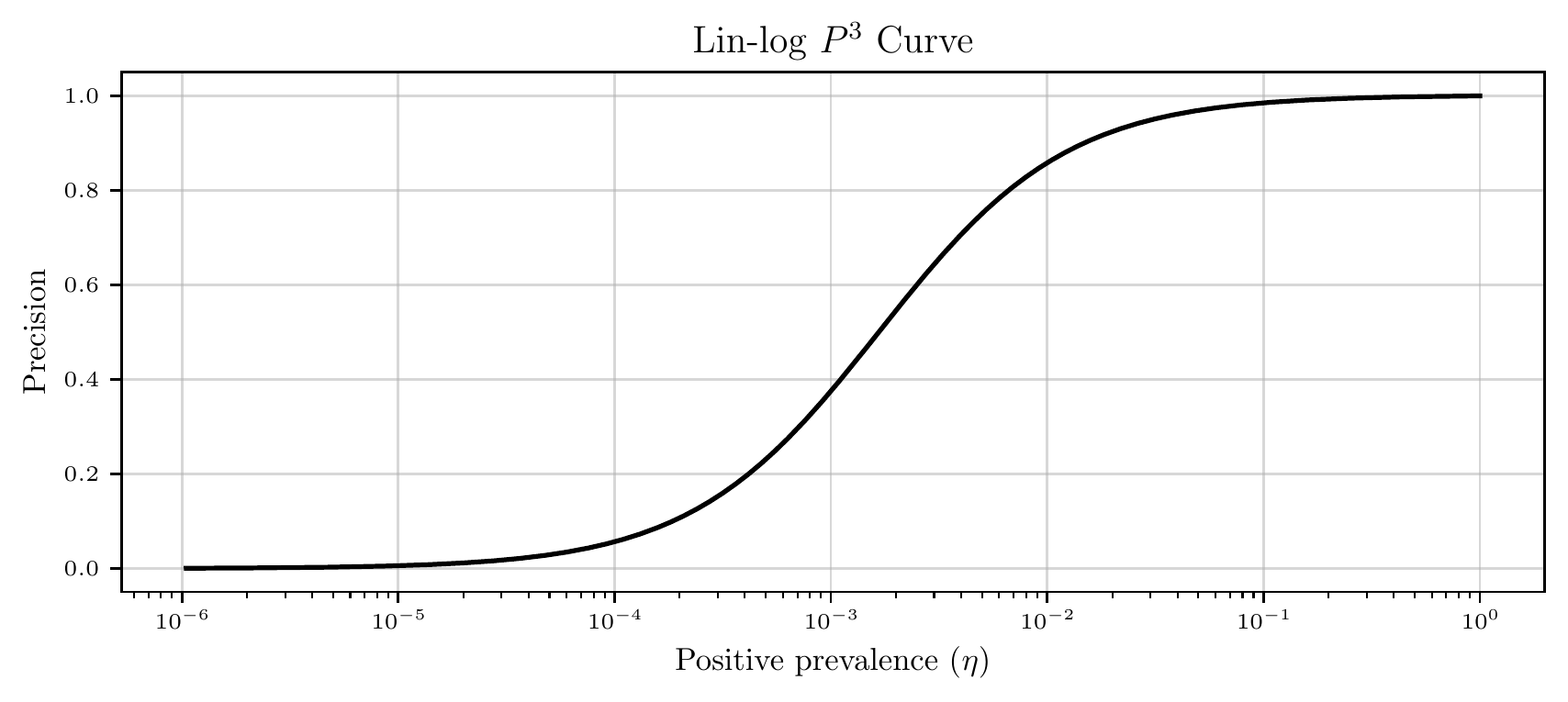} }
    \caption{Positive-Prevalence Precision (\Pthree) curve for a hypothetical classifier with TPR~=~0.6 and FPR = 0.001. The graph is plotted in logarithmic scale of the x-axis.}
    \label{fig:p3}
\end{figure}

% also maybe add curve with recall given a fixed precision on a roc curve?

Positive prevalence adjusted precision computed by Equation \eqref{eq:test_prec} is a linear rational function of the positive class prevalence $\eta$. As such, it can be plotted over an interval of positive prevalence values. We call such plot Positive-Prevalence Precision (\Pthree) curve. The curve should be plotted with log-scaled x-axis (lin-log \Pthree curve) to easily distinguish between different orders of magnitude of the positive prevalence as demonstrated in Figure~\ref{fig:p3}.

\Pthree curve is a useful instrument when evaluating a classifier to determine it's performance beyond a particular dataset. The downside of the plot is that contrary to ROC or PR curves, it captures the performance only for a single operating point of the classifier. Each point on an ROC curve thus has it's corresponding \Pthree curve.

Given a particular ROC curve, each point on the curve corresponds to a different value of TPR. Instead of saying that \Pthree curve corresponds to a particular point on the ROC curve, it can also be said that it corresponds to a fixed value of TPR. For example, \Pthree curve in Figure \ref{fig:p3} corresponds to a classifier with TPR fixed at 60\%.

\Pthree curve answers the question ``How does precision of a given classifier evolve when changing the class imbalance-ratio?'' and allows to quickly visually assess some of the conditions under which the classifier is suitable for production environment. Also, even if \Pthree curve may not be used in a particular evaluation of a classifier it is still important to possess intuition about it's general shape. %Often, in production systems, there are hard constraints on the precision of the classifier. For example, it might be desired that a network intrusion detection system has always precision at least 90\% in order for an analyst to be not overwhelmed by false alarms. Here, the question to be asked is: ``What will be the recall of the system for different imbalance-ratios with precision fixed at a reasonable value?''. This question is addressed in Section \ref{sec:minppx}.

%Include PP-Recall curve for given precision ( plotted from AUC)? Is included in the original paper?

\subsection{Precision-Recall curves}
\label{sec:pr}
PR curve is a very popular method to evaluate classifiers on imbalanced datasets. It captures the relationship between recall (TPR) on the x-axis and precision on the y-axis. As is the case with ROC curve, PR curve is usually created by applying different thresholds on the raw output of a classifier. While ROC is a strictly increasing function, PR curves do not have to be monotonous because it is possible for precision to both increase or decrease for different threshold values. %Include the previous sentence is it necessary?

\begin{figure}[t!]%
    \centering
    {\includegraphics[width=0.45\linewidth]{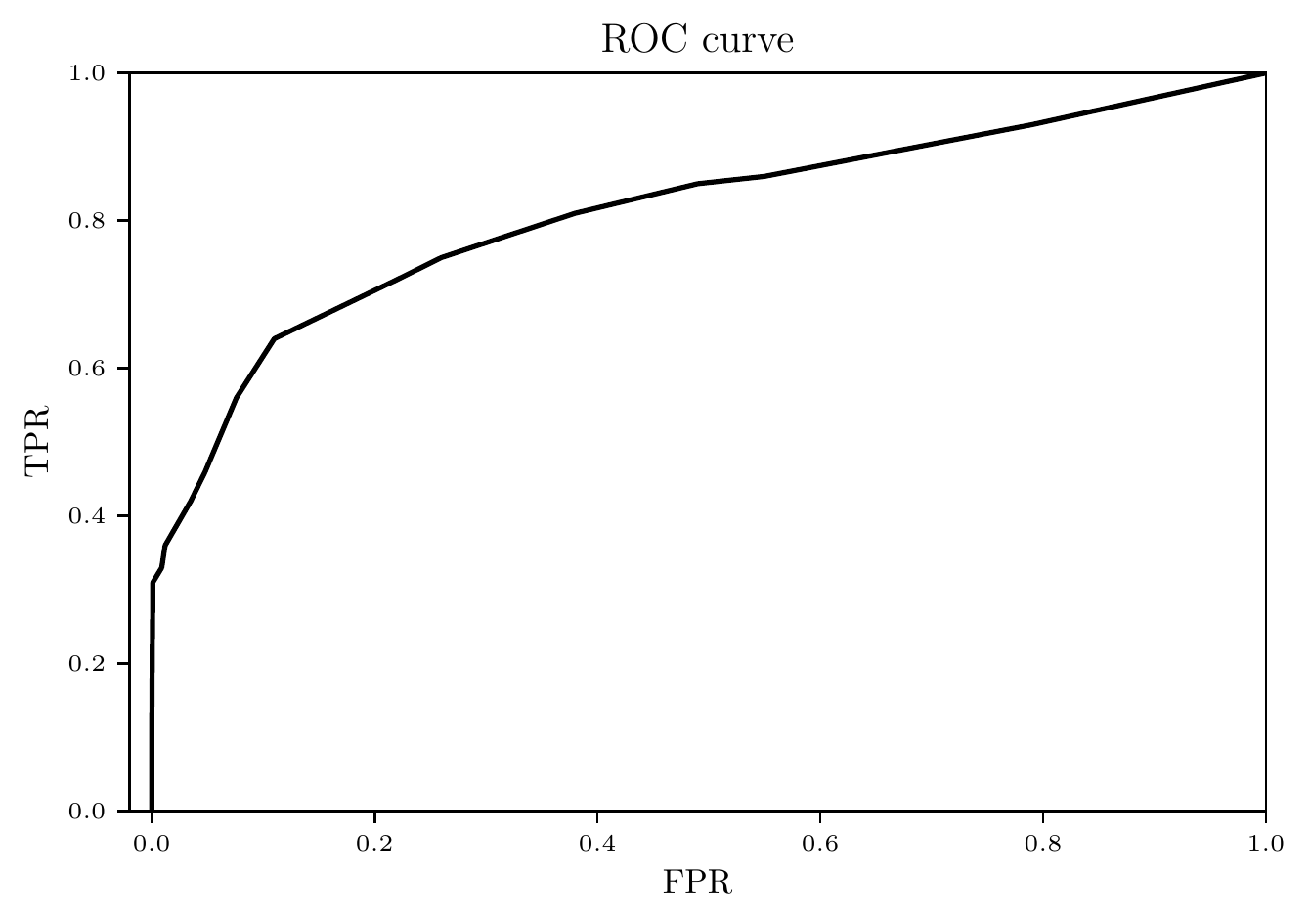} }%
    {\includegraphics[width=0.45\linewidth]{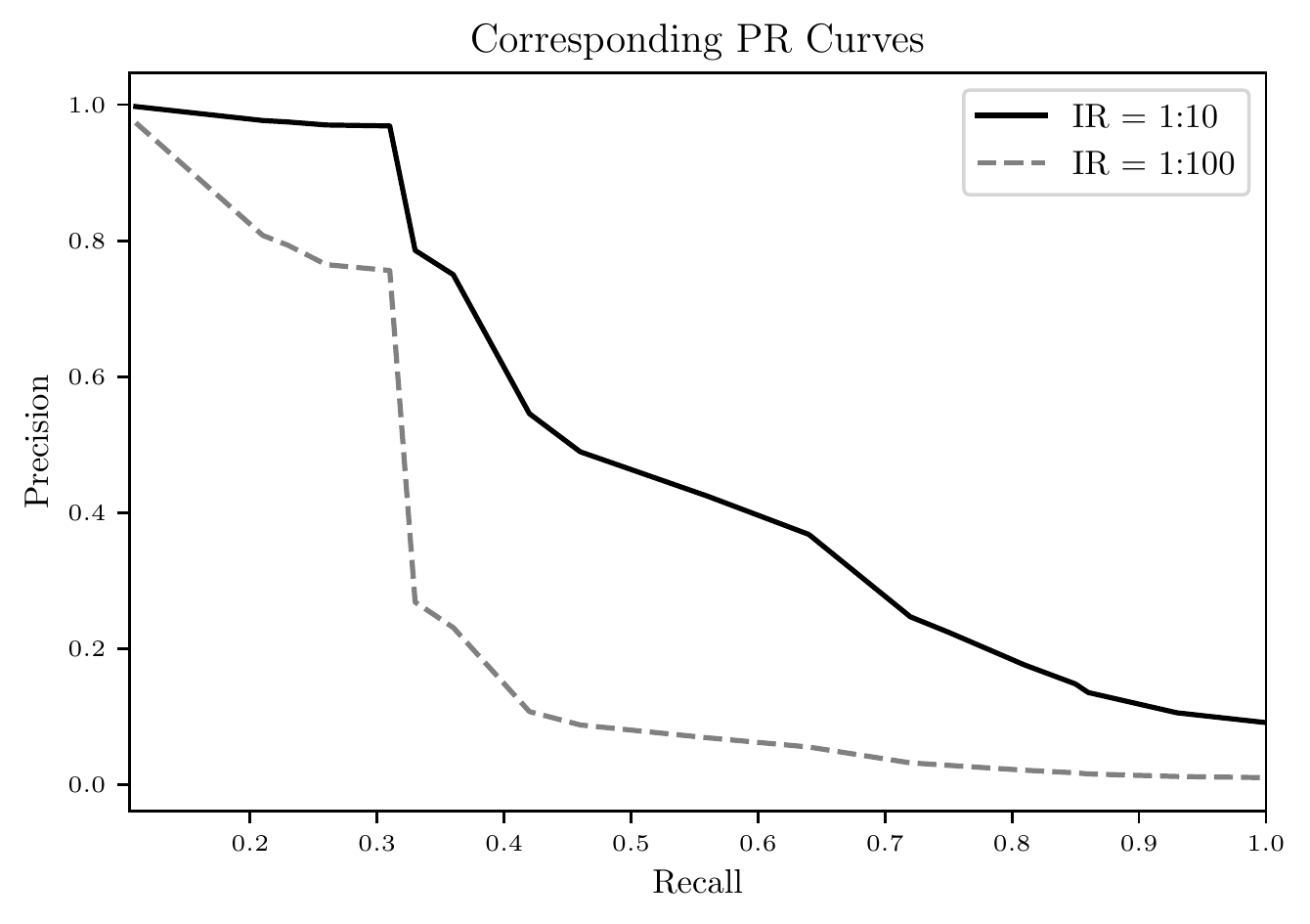} }%
    \caption{Example of how a single ROC curve can correspond to two different PR curves given different imbalance ratios. The solid PR curve was created from the ROC curve with assumption that the IR was $1:10$ while the dashed PR curve corresponds to IR equal to $1:100$.}%
    \label{fig:roc_vs_pr}%
\end{figure}

% dependent on imbalance ratio
As discussed in Section \ref{sec:p3}, contrary to the ROC curve, {\it PR curve is affected by the imbalance ratio present in the test dataset.} This behavior is demonstrated in Figure \ref{fig:roc_vs_pr}. PR curves can immediately reveal poor performance on class-imbalanced datasets that might not be obvious when inspecting ROC curves alone \cite{saito2015precision}. Because of this property PR curves are well suited and popular choice for evaluation of classifiers on class-imbalanced sets.

{\it We suggest that the particular imbalance ratio present in a test dataset for which the PR curve was created should always be reported and considered when interpreting the impact of the results}. When different research teams perform their experiments on different test sets while solving the same problem, and even if the data originate from the same source, the resulting PR curves will not be comparable if different imbalance ratios are present. For example, in computer security the datasets of downloaded files might originate from the VirusTotal\footnote{\url{https://www.virustotal.com}} service, but different teams may work with different subsets that have different imbalance ratios.

Another danger is that the class imbalance ratio in a particular test dataset is often not representative of the imbalance ratios encountered once the classifier is deployed in real environment. It is often the case that the imbalance ratios experienced in the wild are lower than the ratio in the test dataset (not rarely the test datasets are even not imbalanced at all). In such situations, too optimistic estimates of the classifier's performance will be obtained if evaluation based on PR curve computed directly on the test dataset is used.

To remedy these risks, often test datasets with the same class imbalance ratios that would be encountered in the real environment are created. In Section \ref{sec:errors} we demonstrate that this should not be the goal. Rather, a test dataset should be assembled that allows estimation of TPR and FPR with low enough variance and \eqref{eq:test_prec} should be used to compute Precision-Recall curves for different class imbalance ratios of interest. 

% concerned only with positive class - potentially useful paragraph in different section
%Using PR curves alone may be seen as not considering the impact of the classifier on the negative class. Therefore, using PR curves is not suitable in situations where the positive prediction itself negatively affects the members of the negative class. For example, in airport security screening passengers that are identified as positive might have to go through time consuming inspections. In financial fraud detection the result of a positive detection might be that the system blocks customer's credit card. In these cases FPR is of interest because it estimates the fraction of negative class' members that will have to bear the costs of positive prediction. 

\section{Comparing performance of classifiers}
\label{sec:comparing}

When comparing performance of classifiers that need to deal with imbalanced data, the area under PR-curve (PR-AUC) or F1 score ($F_1 = 2 \cdot \frac{\text{Prec}\cdot \text{Recall}}{\text{Prec} + \text{Recall}}$) are often used out of convenience because they can be expressed as a single number \cite{imbalanced_review}. In this section, we show that not only the values of these metrics dramatically depend on the imbalance rate in the selected test dataset, but the rate has notable influence even on the order of classifiers related to their efficacy. That is, based on these metrics two classifiers can switch places given different imbalance rates. This can lead to incorrect conclusions about performance of classifiers on real data. The fact can be also misused for cherry-picking of an imbalance rate to pick the one where a classifier achieves better results than any other method it competes with.

\subsection{Affecting ordering of classifiers: F1 score}
\label{sec:f1}
\begin{figure}[t]
    \centering
    {\includegraphics[width=0.8\linewidth]{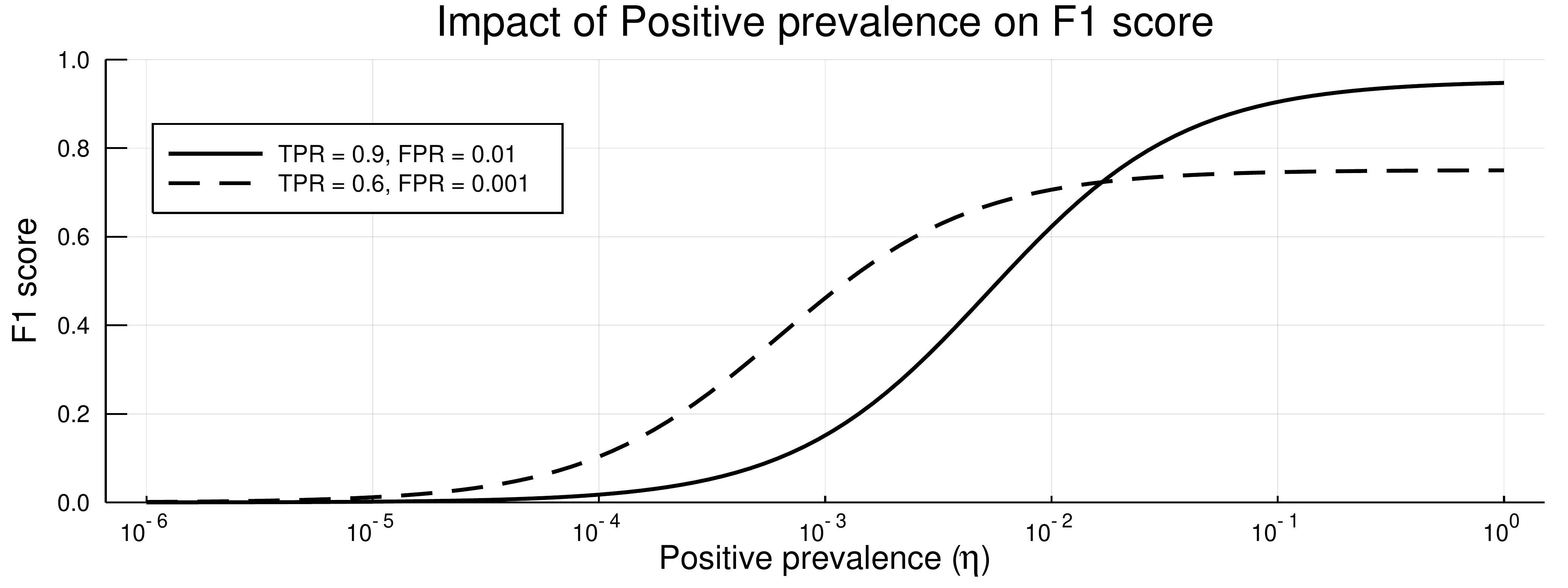}}
    \caption{The graph is similar to Positive Prevalence-Precision plot in Figure \ref{fig:p3} but instead of precision it plots F1 score of two distinct classifiers computed on the same dataset but assuming different imbalance rates. It can be seen that not only the absolute value of the score but even the order of the classifiers depends on the positive class prevalence.}
    \label{fig:fscorepp}
\end{figure}

F1 score is defined as harmonic mean of precision and recall. The comparison of F1 scores of two classifiers is therefore affected by the selected imbalanced rate since precision depends on the rate while recall does not. Figure~\ref{fig:fscorepp} demonstrates how the F1 score of two classifiers depends on the imbalance rate present in a test dataset.

Therefore, {\it we suggest to plot F1 scores in relation to imbalance rates, such as seen in Figure \ref{fig:fscorepp} instead of tabulated F1 scores in any applied research papers}. The plot contains a superset of information, it is easily interpretable, space-efficient and conveys an overall better picture about performance of classifiers independent of the particular imbalance rate in the selected test dataset. The imbalance rate of the particular test dataset can be easily highlighted on the x-axis.

\subsection{Affecting ordering of classifiers: PR-AUC}

Firstly, it is proven that if a classifier {\it dominates} in ROC space it also dominates in PR space \cite{davis2006relationship}, but dominance is not linked to the area under ROC curve (ROC-AUC). It is easily possible for a classifier to have greater ROC-AUC than another but smaller area under PR curve (PR-AUC) on the same test dataset.

A convenient property of evaluating classifier by ROC-AUC is that it's value is invariant to class imbalance. On the other hand, the value of ROC-AUC can be dominated by insignificant regions in the ROC space, e.g. high values of FPR, which are in practice of no importance. If the problem is heavily class imbalanced it is usually not an appropriate method for evaluation of classifiers~\cite{bm-eval} and PR-AUC should be considered.

\begin{figure}[t!] %TODO: Name the axes, decide if we want to include all of them
    \centering
    {\includegraphics[width=0.44\linewidth]{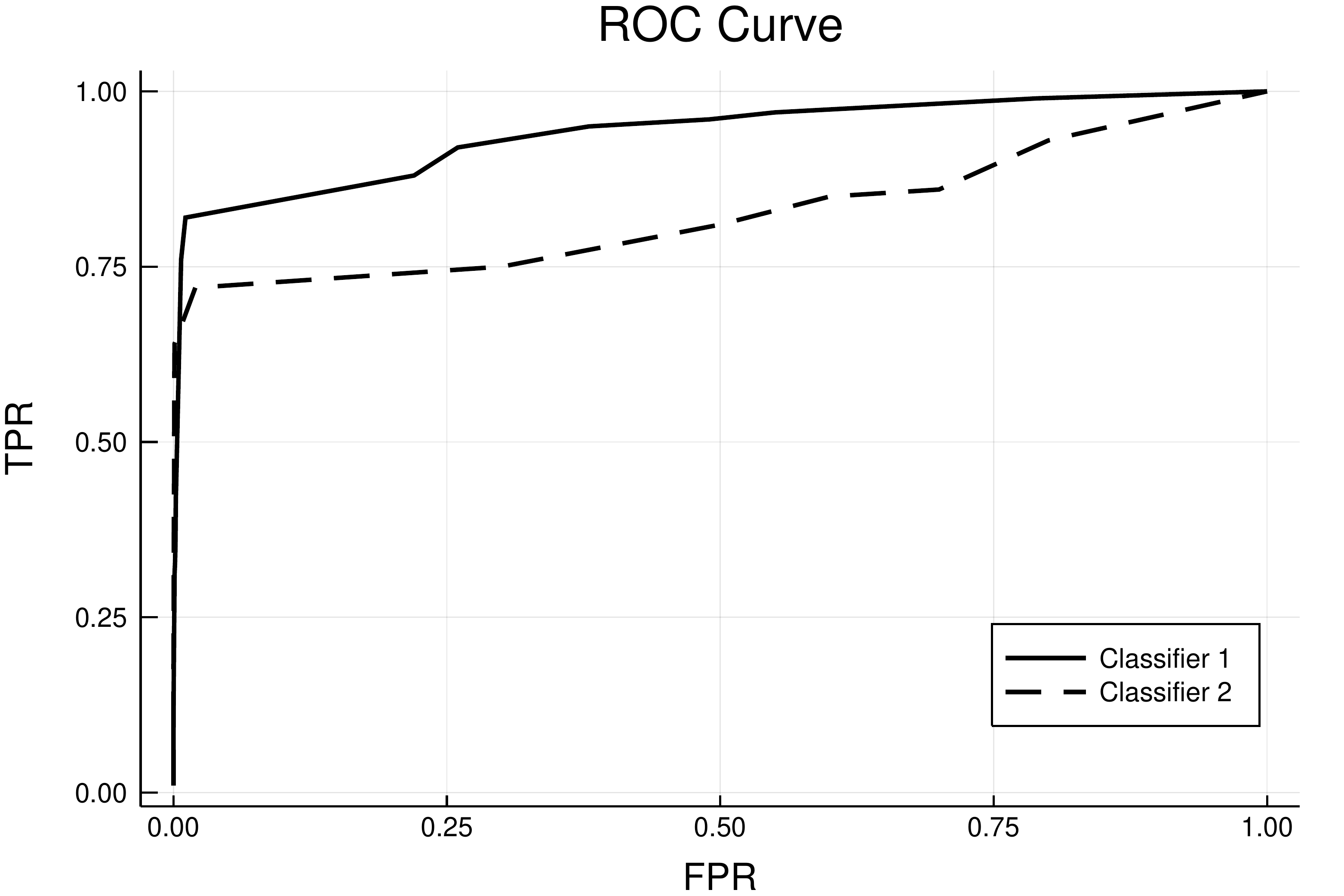}}
    {\includegraphics[width=0.44\linewidth]{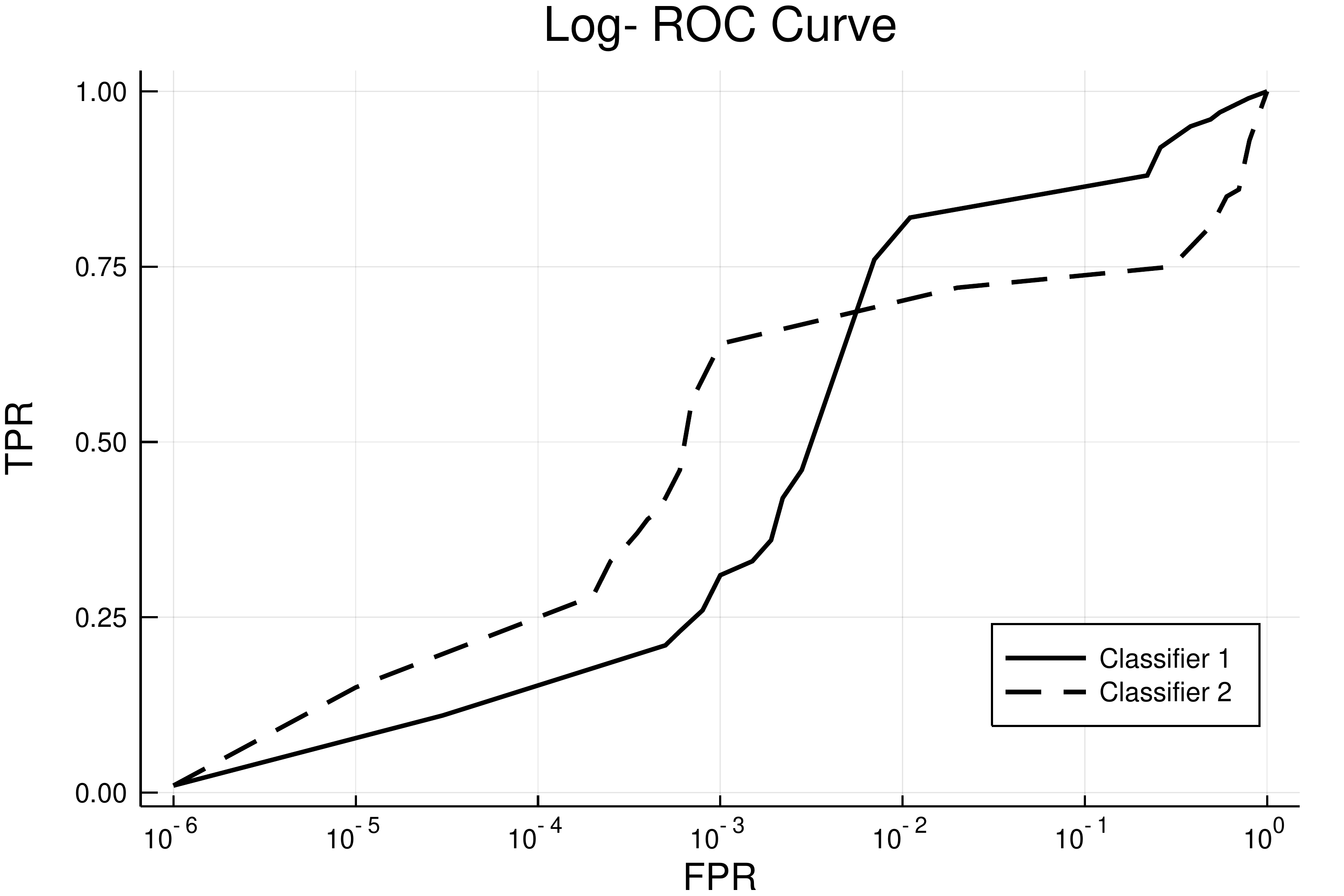}}
    {\includegraphics[width=0.44\linewidth]{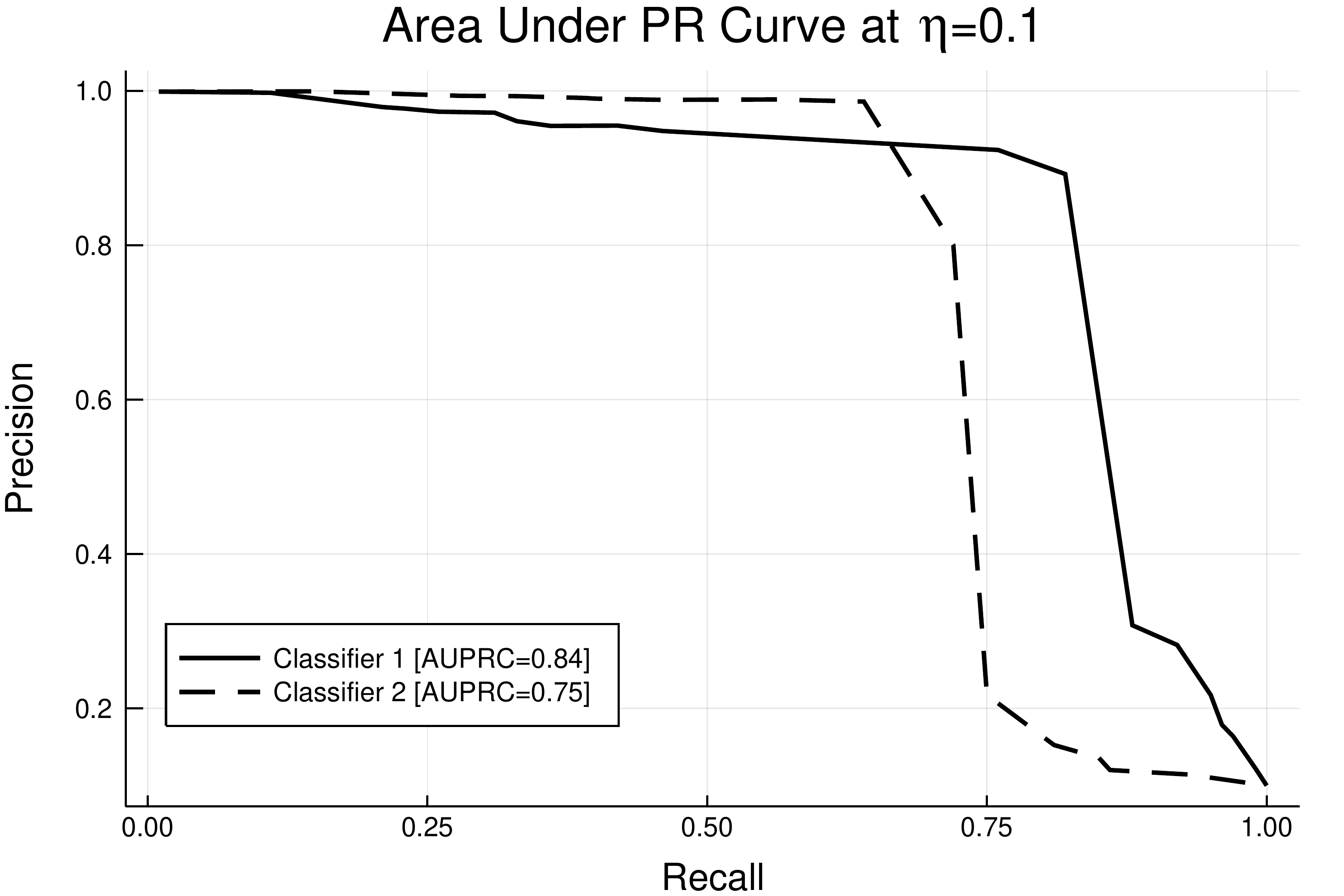}}
    {\includegraphics[width=0.44\linewidth]{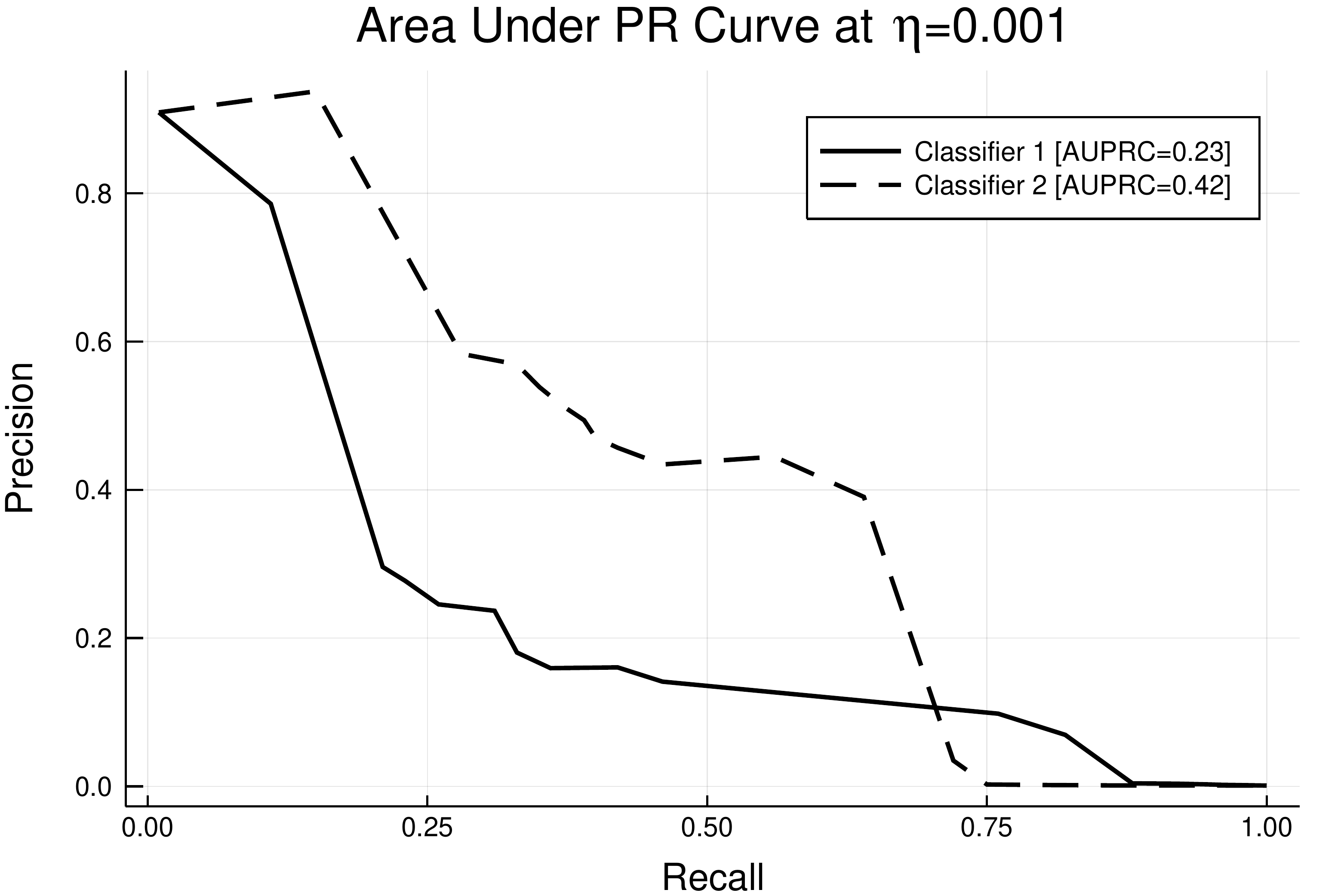}}
    {\includegraphics[width=0.88\linewidth]{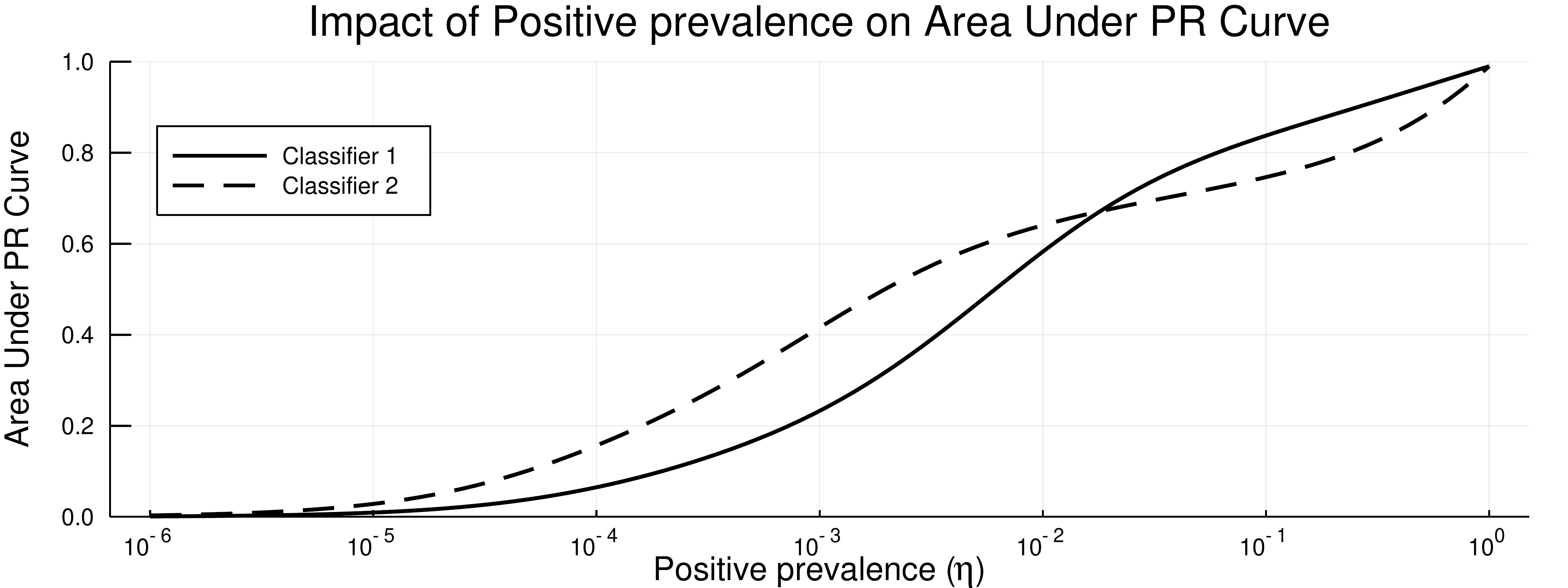}}
    \caption{The top-left plot is an example plot of two classifier ROC curves. In the top-right plot the same ROC curves are displayed with logarithmically scaled x-axis. The middle row displays corresponding PR curves for the ROC curves under different positive class prevalences (namely $10^{-1}$ and $10^{-3}$). The bottom plot shows how PR-AUC of the classifiers depends on the class imbalance rate and that the order of the classifiers can easily switch for two different prevalences.}
    \label{fig:prauc}
\end{figure}

However, it is often not realized that PR-AUC values depend on class imbalance and notably that also the order of classifiers under this metric depends on the imbalance rate as demonstrated in Figure \ref{fig:prauc}. It may be more surprising than in the case of F1 score computed only at a single operating point, while PR-AUC is evaluated over the whole range of operating points. Therefore, one might wrongly expect the metric to preserve ordering of classifiers across different imbalance rates.

We offer similar advice as with F1 score about the need to report the dataset imbalance rate together with PR-AUC values and to ideally use plots as in Figure \ref{fig:prauc} instead of tabulated values for a single imbalance ratio.

%To extend the results to other imbalance rates, \eqref{eq:prec} can be used to transform PR curve to different imbalance rate and then compute PR-AUC. We strongly suggest to use plot of PR-AUC with respect to the positive class prevalence as shown in Figure \ref{fig:prauc} instead of just tabulating the PR-AUC results on the particular test dataset. The rationale for this are the same as in the case of F1 score mentioned in Section \ref{sec:f1}.

\section{Impact of errors on estimates of TPR and FPR } %Evaluation metrics measurement errors in class-imbalanced problems}
\label{sec:errors}

Class-imbalanced problems have increased demands on the test dataset size. It is often ignored that $\tTPR$ and $\tFPR$ computed on test dataset are just point estimates of the real TPR and FPR, given in \eqref{eq:tpr} and \eqref{eq:fpr}, respectively, and as such they may be affected by uncertainty related to insufficient amount of samples of the minority class.
In this section, we investigate how this uncertainty impacts the measured precision and how to correctly design experiments in presence of imbalanced data to suppress the uncertainty in the outcome.

%The point estimates $\tTPR$ and $\tFPR$ computed by \eqref{eq:tpr} and \eqref{eq:fpr} are realizations of random variables. If the size of the test dataset is small the distributions of the random variables is flat and the estimates are not reliable as well as the \Pthree curve derived from them.

%\subsection{Coefficient of variation}
%\label{sec:cv}

A common approach to quantify the uncertainty of estimates based on finite samples is to use the interval estimates. We say that $\ITpr=(\tTPR-\SigmaTpr,\tTPR+\SigmaTpr)$ is the $\alpha$-confidence interval of $\TPR$ if it holds that
\begin{equation}
\Prob( \TPR \in \ITpr )\geq \alpha \,,
\end{equation}
where the probability is w.r.t. randomly generated positive test samples $\mathcal{X}^+$ which are used to compute $\tTPR$ by~\eqref{equ:estimatedTPR}. The interval (half-)width $\SigmaTpr$, the number of samples $|\mathcal{X}^+|$ and the confidence level $\alpha \in (0,1)$ are dependent variables the exact relation of which is characterized by numerous concentration bounds like the Hoeffding's inequality. For example, by fixing $\SigmaTpr$ and $\alpha$ we can compute the minimal number of samples in $\mathcal{X}^+$ which guarantee that $\ITpr$ is the $\alpha$-confidence interval. In the sequel we assume that the interval width $\SigmaTpr$ is not greater than $\tTPR$. Note that this formalisation does not introduce any specific constraints on the shape of $\TPR$ distribution. The confidence interval $\ITpr$ can be characterized by a single number, the coefficient of variation, defined as 
\begin{equation}
   \CTpr = \frac{\SigmaTpr}{\tTPR}\:.
\end{equation}
Analogously, we can define $\IFpr=(\tFPR-\SigmaFpr,\tFPR+\SigmaFpr)$, $\CFpr=\frac{\SigmaFpr}{\FPR}$, and we also assume that $\SigmaFpr< \tFPR$.

\begin{figure}[t]
    \centering
    \includegraphics[width=1\textwidth]{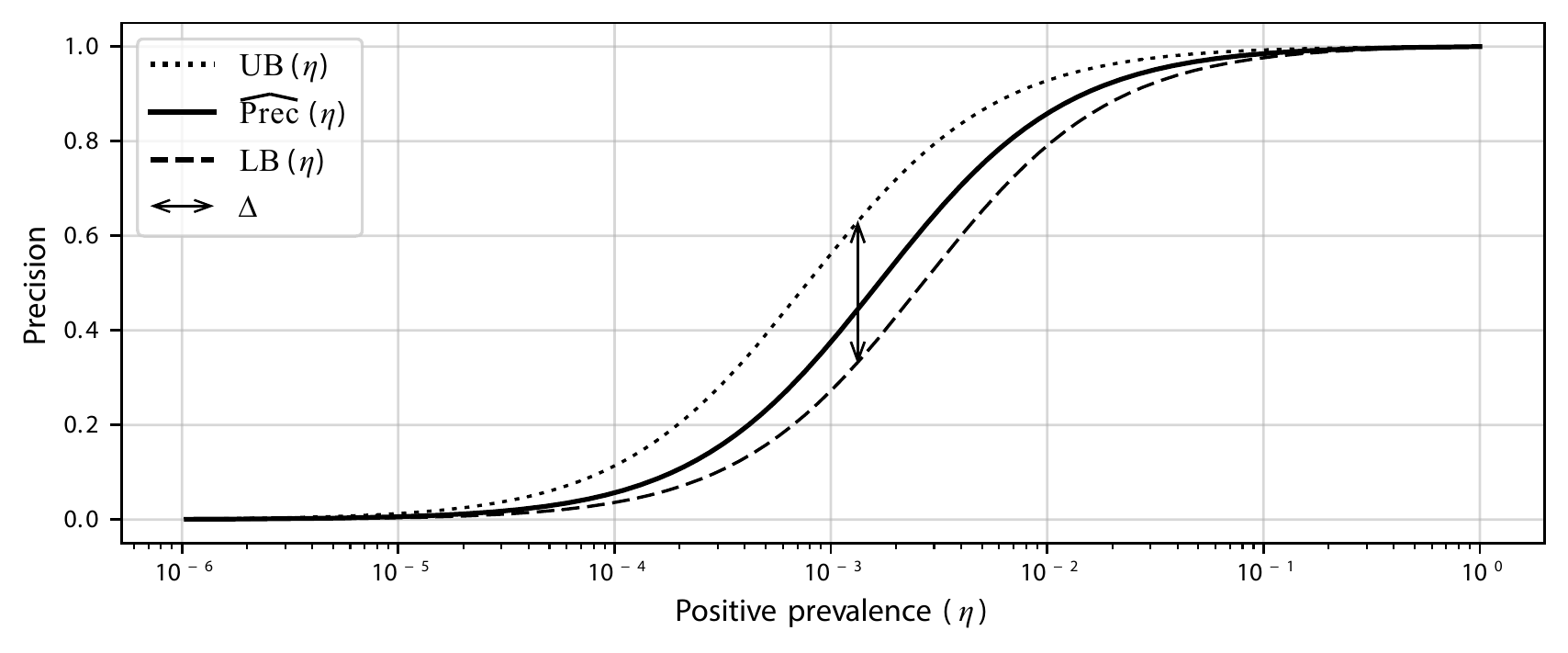}
    \caption{The figure visualizes the uncertainty band containing the value of $\Prec(\eta,\TPR,\FPR)\in (\UB(\eta),\LB(\eta))$ when $\TPR$ and $\FPR$ are bound to intervals $\ITpr=(\tTPR-\SigmaTpr,\tTPR+\SigmaTpr)$ and $\IFpr=(\tFPR-\SigmaFpr,\tFPR+\SigmaFpr)$, respectively. The value $\Delta=\max_{\eta\in(0,1)}(\UB(\eta)-\LB(\eta))$
    corresponds to the maximal width of the uncertainty band. The solid line corresponds to the point estimate $\tPrec(\eta)=\Prec(\eta,\tTPR,\tFPR)$. }
    \label{fig:PrecBand}
\end{figure}

Let us define the precision as a function of the positive class prevalence $\eta$, $\TPR$ and $\FPR$~\footnote{In \eqref{eq:test_prec} we used $\Prec(\eta)$ since the values of $\TPR$ and $\FPR$ were assumed to be fixed.}:
\begin{equation}
  \Prec(\eta,\TPR,\FPR) = \frac{\eta\cdot\TPR}{\eta\cdot\TPR+(1-\eta)\cdot \FPR} \:.
\end{equation}
Given $\TPR\in\ITpr$ and $\FPR\in\IFpr$, the value of $\Prec(\eta,\TPR,\FPR)$ has to be for any fixed $\eta\in(0,1)$ inside the interval $(\LB(\eta),\UB(\eta))$ where
\begin{eqnarray}
\label{equ:lb}
 \displaystyle \LB(\eta) = \min_{\TPR\in\ITpr\atop{\FPR\in\IFpr}} \Prec(\eta,\TPR,\FPR)\,,\\
 \label{equ:ub}
\displaystyle \UB(\eta) = \max_{\TPR\in\ITpr\atop{\FPR\in\IFpr}} \Prec(\eta,\TPR,\FPR) \:.
\end{eqnarray}
Let $\Delta$ be the maximal width of the interval $(\LB(\eta),\UB(\eta))$ w.r.t. $\eta$, that is,
\begin{equation}
  \Delta = \max_{\eta\in(0,1)}(\UB(\eta)-\LB(\eta))\:.
\end{equation}
The number $\Delta$ can be interpreted as the maximal uncertainty in measurements of precision when the exact values of $\TPR$ and $\FPR$ are replaced by their confidence intervals $\ITpr$ and $\IFpr$, respectively. It is easy to see that $\TPR\in\ITpr$ and $\FPR\in\IFpr$ imply
\begin{equation}
  \Prec(\eta,\TPR,\FPR) \in (\tPrec(\eta)-\Delta,\tPrec(\eta)+\Delta) \:.
\end{equation}
The concepts of $\UB(\eta)$, $\LB(\eta)$ and $\Delta$ as well as their relation to $\Prec(\eta,\TPR,\FPR)$ are illustrated in Figure~\ref{fig:PrecBand}. The following theorem relates the maximal uncertainty $\Delta$ and the coefficients of variation $\CTpr$ and $\CFpr$, which characterize the confidence intervals $\ITpr$ and $\IFpr$, respectively. 

\begin{theorem}
\label{th:delta_cv}
  Let $\TPR\in(\tTPR-\SigmaTpr,\tTPR+\SigmaTpr)$ and $\FPR\in(\tFPR-\SigmaFpr,\tFPR+\SigmaFpr)$. Let further $\tTPR>\SigmaTpr$ and $\tFPR>\SigmaFpr$. Then 
  \[
      \Delta \leq \max\{ \CTpr, \CFpr\}
  \]
  and the equality is attained iff $\CTpr = \CFpr$.\footnote{The proof for Theorem \ref{th:delta_cv} is available in the Section \ref{sec:th1proof} of appendix.}
\end{theorem}

\begin{corollary}
  Let $\ITpr$ and $\IFpr$ be $\alpha$-confidence intervals of the true $\TPR$ and $\FPR$, respectively, and let $\CTpr$ and $\CFpr$ be their corresponding coefficients of variation. Let further $\Delta = \max\{ \CTpr, \CFpr\}$.
  Then $\IPrec=(\tPrec(\eta)-\Delta,\tPrec(\eta)+\Delta)$ is the $\alpha^2$-confidence interval of $\Prec(\eta,\TPR,\FPR)$, i.e.
  \[
  \Prec(\eta,\TPR,\FPR) \in \IPrec 
 \]
 holds with probability $\alpha^2$ at least. 
\end{corollary}
% should Delta or UB(\eta) - LB(\eta) be here? or should it be reformulated so it holds with probability at least alpha^2?
% refer to proof for theorem 1

The $\alpha^2$-confidence level stems from the fact that $\TPR\in\ITpr$ and $\FPR\in\IFpr$ are two independent random events with probability not less than $\alpha$.

Theorem~\ref{th:delta_cv} shows the relationship between confidence intervals for precision, widths of these intervals and point estimates of TPR, FPR. That is, coefficients of variation for TPR and FPR are the crucial quantities to consider when designing test dataset. If a test set is constructed we first need to manually fix both $\SigmaTpr$ and $\SigmaFpr$ at reasonable values based on the purpose of the dataset, and then ensure sufficient number of testing samples necessary to estimate TPR, FPR with desired $\Delta$. If, for example, one is interested in FPR $=10^{-3}$ on a dataset having only 10,000 negative samples, the estimate around this working point may become extremely noisy. Since such low FPR corresponds to only 10 FP samples (10,000 * $10^{-3}$), just a small increase or decrease in number of FPs suffice to significantly alter the relative value of the FPR. Therefore, if such low values of FPR are of interest, one should increase the amount of negatives. Different methods exist that can quantify the concentration bounds. For example, Hoeffding's inequality can be used, which states that the upper bound on the number of required samples is proportional to $\frac{1}{\SigmaFpr^2}$, but Hoeffding's bound is very loose and usually less samples are required. 

On the other hand, given a test dataset, in order to find $\Delta$ we need to estimate $\SigmaTpr, \SigmaFpr$ to get $\CTpr, \CFpr$. For that purpose cross-validation or bootstrapping can be used. For example, a classifier with $\tTPR = 0.6, \SigmaTpr = 0.06, \tFPR = 10^{-3}, \SigmaFpr = 10^{-4}$ has $\CTpr = \CFpr = \Delta = 0.1$, which might be reasonable width of the precision's confidence interval (i.e. $\pm 10\%$ change). But, if we increase $\SigmaFpr = 5*10^{-4}$ then even though the number might seem small and it may be not indicative of the impact on estimate of the precision, the bound for precision becomes $\Delta=0.5$ (i.e. $\pm 50\%$ change), which will immediately shed light on the reliability of estimates of the precision.\footnote{In this example, $\Delta \approx 0.31$ for $\eta \approx 1.45\cdot10^{-3}$. Computation can be found in the supplementary code to this paper.}

\subsection{Example of errors caused by sub-sampling} % include 'experiment' in name so reviewers do not miss it

% \cite{russakovsky2015imagenet}, \cite{He_2016_CVPR}

\begin{figure}[t]
    \centering
    \includegraphics[width=1\textwidth]{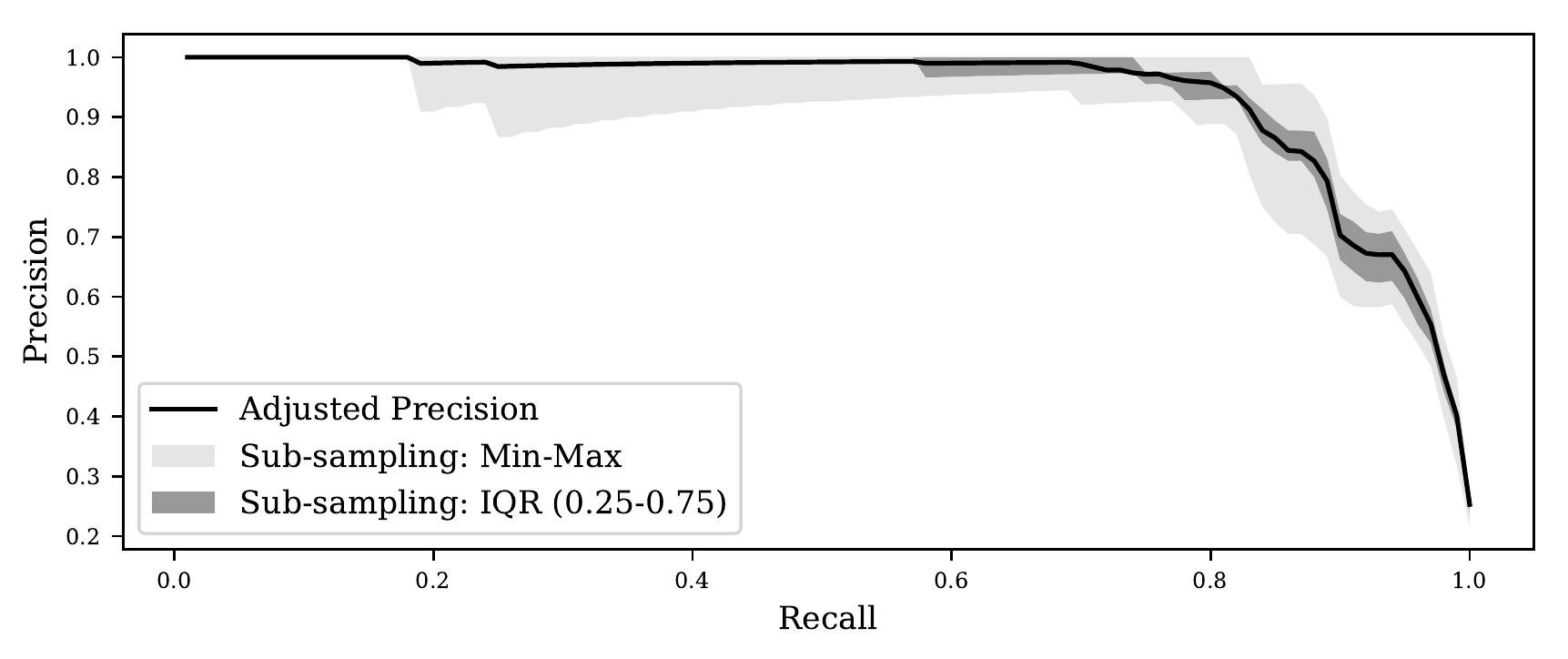}
    \caption{PR curves for $\eta = 10^{-2}$. The black PR curve is computed from full dataset with $p_{+}=10^{-3}$ and adjusted to $\eta = 10^{-2}$ using \eqref{eq:test_prec}, whereas the gray areas indicate IQR and min-max range of PR curves computed on 30 datasets with randomly sub-sampled negative class to match $p_{+}=10^{-2}$. Note that some PR curves are inside of IQR only partially.}
    \label{fig:imagenet}
\end{figure}

To illustrate the error of sub-sampling we used ResNet-50 \cite{He_2016_CVPR} on the ImageNet validation dataset \cite{russakovsky2015imagenet} to detect images of 'agama' in a one-vs-all manner. The $p_{+}$ in such dataset is $10^{-3}$.

To plot PR curves for $\eta = 10^{-2}$ we can either use the full dataset and then apply \eqref{eq:test_prec} to adjust the precision, or sub-sample the dataset to $p_{+} = 10^{-2}$. Figure \ref{fig:imagenet} compares these two approaches, where we repeated the sub-sampling 30 times to estimate the variance introduced by random reduction of the negative class. The results show that PR curves measured on the sub-sampled datasets are encumbered by a considerable measurement errors even though each one has 5000 samples, which might otherwise be a reasonable number for evaluation on balanced problems. Moreover $\eta = 10^{-2}$ is not as drastic imbalance as is often encountered in applications and the errors could be even more pronounced if $\eta$ was lower.

Unlike the common practice of sub-sampling of the test dataset to the desired imbalance rate \cite{pendlebury2019tesseract}, we recommend to use a bigger dataset (to decrease the coefficients of variation) and adjust the metrics to the desired imbalance rate instead.

\section{Related Work}

Several comprehensive papers about methodology of evaluation on imbalanced datasets were written \cite{chawla2009data,fawcett2006introduction,he2009learning,kotsiantis2006handling,sokolova2009systematic}. They focus on measuring the performance on the test dataset and do not address the problem of mismatch between class imbalances in test and application datasets.

In \cite{damodaran2017comparison} authors use a plot with area under PR curve on the y-axis and a quantity related to the imbalance ratio on the x-axis. The plot is similar to Figure~\ref{fig:prauc}, it is used because it is useful in the context of the paper but it's properties and impacts are not discussed.

In \cite{bm-eval} authors discuss several bad practices in handling of class-imbalanced problems. Apart from other causes, they discuss the importance of addressing the real imbalance ratios that can be different from the test dataset. They also present a formula for adjusting the precision to different imbalance ratios but do not explore this formula in greater detail neither inspect the impact of uncertainty originating from the finite size of the test dataset on precision.

Paper \cite{landgrebe2006precision} introduces measure based on area under PR curve, which is further integrated across different class imbalances yielding a single evaluation number. The idea is based on the relationship between PR and ROC given in \eqref{eq:test_prec}. No additional investigations related to multiple working points, ordering of classifiers according to the score, nor errors in measurements are carried out.

In \cite{pendlebury2019tesseract} authors raise the issue of experimental results in cybersecurity often not being reproducible in real applications. They mention the problem that the class imbalance is often different in test dataset and in practice. They do not address the issue analytically but instead choose to re-sample the test dataset to desired imbalance ratios. This goes directly against our observations in Section \ref{sec:errors} and applying such method leads to results heavily affected by noise.

It should be mentioned that other evaluation metrics well-suited for evaluation of class-imbalanced problems were proposed. A notable example is Matthews Correlation Coefficient (MCC) \cite{matthews1975comparison}, but is not in the scope of this paper. MCC is not as widely used as PR \cite{imbalanced_review} and it's values are not that easily interpretable as values of precision and recall.

\section{Conclusion}

This paper addressed evaluation of classifiers under consideration that the class imbalance ratio encountered in real world is different from imbalance present in the test dataset or is suspect to change. %This situation arises frequently and across many application domains.
We focused on precision as one of the most popular evaluation metrics for imbalanced problems. 

%It is not necessary to obtain a test dataset that is identically distributed as the real-world application because the precision (and metrics that are based on precision) can be adjusted to the desired class imbalance easily and also precision of a classifier can be investigated across different imbalance ratios with the \Pthree plot.

%We have shown how to investigate the performance of a classifier in such situations. More concretely, we focused on the precision metric, because it's result depends heavily on the class imbalance ratio. It is not necessary to obtain a test dataset that is identically distributed as the real-world application because the precision (and metrics that are based on precision) can be adjusted to the desired class imbalance easily and also precision of a classifier can be investigated across different imbalance ratios with the \Pthree plot.
We stress that it is of significant importance to report also the imbalance ratio under which the classifier was developed and is aimed for, because assuming different imbalance ratios may easily lead to swapping of places of classifiers. This holds also for both PR-AUC and F1 score.
%Considering correct imbalance ratios is important because not only the absolute values of commonly used metrics such as F1 score and area under PR Curve (PR-AUC) depend on it, but also, interestingly, the order of the classifiers can switch at different imbalances ratios. This can result in incorrect decisions about which classifier is best suited for a task if they are evaluated on a different class imbalance rate. Graphs relating F1 score or PR-AUC to imbalance rate should be preferred to just tabulating the results, because they contain a superset of information.

We have shown that even very small absolute values of $\sigma_{\text{FPR}}$ can result in large variance in measured precision. The larger the class imbalance, the greater are the demands on the amount of negative samples present in the test dataset. Therefore, rather than sub-sampling a dataset to reach desired imbalance rate, all the samples should be kept to decrease the coefficients of variation, and the evaluation metrics should be computed given the presented formulas.

 \bibliographystyle{splncs04}
 \bibliography{main}

\begin{thebibliography}{10}
\providecommand{\url}[1]{\texttt{#1}}
\providecommand{\urlprefix}{URL }
\providecommand{\doi}[1]{https://doi.org/#1}

\bibitem{axelsson2006base}
Axelsson, S., Sands, D.: The base-rate fallacy and the difficulty of intrusion
  detection. Understanding Intrusion Detection Through Visualization pp. 31--47
  (2006)

\bibitem{bm-eval}
Brabec, J., Machlica, L.: Bad practices in evaluation methodology relevant to
  class-imbalanced problems. Critiquing and Correcting Trends in Machine
  Learning workshop at NeurIPS  \textbf{abs/1812.01388} (2018),
  \url{http://arxiv.org/abs/1812.01388}

\bibitem{bayesianforests}
Brabec, J., Machlica, L.: Decision-forest voting scheme for classification of
  rare classes in network intrusion detection. In: 2018 IEEE International
  Conference on Systems, Man, and Cybernetics (SMC). pp. 3325--3330 (Oct 2018).
  \doi{10.1109/SMC.2018.00563}

\bibitem{chawla2009data}
Chawla, N.V.: Data mining for imbalanced datasets: An overview. In: Data mining
  and knowledge discovery handbook, pp. 875--886. Springer (2009)

\bibitem{damodaran2017comparison}
Damodaran, A., Di~Troia, F., Visaggio, C.A., Austin, T.H., Stamp, M.: A
  comparison of static, dynamic, and hybrid analysis for malware detection.
  Journal of Computer Virology and Hacking Techniques  \textbf{13}(1),  1--12
  (2017)

\bibitem{davis2006relationship}
Davis, J., Goadrich, M.: The relationship between precision-recall and roc
  curves. In: Proceedings of the 23rd international conference on Machine
  learning. pp. 233--240. ACM (2006)

\bibitem{fawcett2006introduction}
Fawcett, T.: An introduction to roc analysis. Pattern recognition letters
  \textbf{27}(8),  861--874 (2006)

\bibitem{imbalanced_review}
Haixiang, G., Yijing, L., Shang, J., Mingyun, G., Yuanyue, H., Bing, G.:
  Learning from class-imbalanced data: Review of methods and applications.
  Expert Systems with Applications  \textbf{73},  220--239 (2017)

\bibitem{he2009learning}
He, H., Garcia, E.A.: Learning from imbalanced data. IEEE Transactions on
  knowledge and data engineering  \textbf{21}(9),  1263--1284 (2009)

\bibitem{He_2016_CVPR}
He, K., Zhang, X., Ren, S., Sun, J.: Deep residual learning for image
  recognition. In: The IEEE Conference on Computer Vision and Pattern
  Recognition (CVPR) (June 2016)

\bibitem{kotsiantis2006handling}
Kotsiantis, S., Kanellopoulos, D., Pintelas, P., et~al.: Handling imbalanced
  datasets: A review. GESTS International Transactions on Computer Science and
  Engineering  \textbf{30}(1),  25--36 (2006)

\bibitem{landgrebe2006precision}
Landgrebe, T.C., Paclik, P., Duin, R.P.: Precision-recall operating
  characteristic (p-roc) curves in imprecise environments. In: Pattern
  Recognition, 2006. ICPR 2006. 18th International Conference on. vol.~4, pp.
  123--127. IEEE (2006)

\bibitem{matthews1975comparison}
Matthews, B.W.: Comparison of the predicted and observed secondary structure of
  t4 phage lysozyme. Biochimica et Biophysica Acta (BBA)-Protein Structure
  \textbf{405}(2),  442--451 (1975)

\bibitem{pendlebury2019tesseract}
Pendlebury, F., Pierazzi, F., Jordaney, R., Kinder, J., Cavallaro, L.:
  $\{$TESSERACT$\}$: Eliminating experimental bias in malware classification
  across space and time. In: 28th $\{$USENIX$\}$ Security Symposium
  ($\{$USENIX$\}$ Security 19). pp. 729--746 (2019)

\bibitem{Phua:2004:MRF:1007730.1007738}
Phua, C., Alahakoon, D., Lee, V.: Minority report in fraud detection:
  Classification of skewed data. SIGKDD Explor. Newsl.  \textbf{6}(1),  50--59
  (Jun 2004). \doi{10.1145/1007730.1007738},
  \url{http://doi.acm.org/10.1145/1007730.1007738}

\bibitem{rahman2013addressing}
Rahman, M.M., Davis, D.: Addressing the class imbalance problem in medical
  datasets. International Journal of Machine Learning and Computing
  \textbf{3}(2), ~224 (2013)

\bibitem{russakovsky2015imagenet}
Russakovsky, O., Deng, J., Su, H., Krause, J., Satheesh, S., Ma, S., Huang, Z.,
  Karpathy, A., Khosla, A., Bernstein, M., et~al.: Imagenet large scale visual
  recognition challenge. International journal of computer vision
  \textbf{115}(3),  211--252 (2015)

\bibitem{saito2015precision}
Saito, T., Rehmsmeier, M.: The precision-recall plot is more informative than
  the roc plot when evaluating binary classifiers on imbalanced datasets. PloS
  one  \textbf{10}(3),  e0118432 (2015)

\bibitem{sokolova2009systematic}
Sokolova, M., Lapalme, G.: A systematic analysis of performance measures for
  classification tasks. Information Processing \& Management  \textbf{45}(4),
  427--437 (2009)

\bibitem{wei2013effective}
Wei, W., Li, J., Cao, L., Ou, Y., Chen, J.: Effective detection of
  sophisticated online banking fraud on extremely imbalanced data. World Wide
  Web  \textbf{16}(4),  449--475 (2013)

\bibitem{yu2010multiscale}
Yu, L., Wang, S., Lai, K.K., Wen, F.: A multiscale neural network learning
  paradigm for financial crisis forecasting. Neurocomputing  \textbf{73}(4-6),
  716--725 (2010)

\end{thebibliography}

\appendixtitleon
%\appendixtitletocon

\begin{subappendices}

\renewcommand{\thesection}{\Alph{section}}

\section{Proof of Theorem \ref{th:delta_cv}}
\label{sec:th1proof}

\newtheorem{innercustomthm}{Theorem}
\newenvironment{customthm}[1]
  {\renewcommand\theinnercustomthm{#1}\innercustomthm}
  {\endinnercustomthm}
  
\begin{customthm}{1} \label{th:doundOnDelta1}
  Let $\TPR\in(\tTPR-\SigmaTpr,\tTPR+\SigmaTpr)$ and $\FPR\in(\tFPR-\SigmaFpr,\tFPR+\SigmaFpr)$. Let further $\tTPR>\SigmaTpr$ and $\tFPR>\SigmaFpr$. Then 
  \[
      \Delta \leq \max\{ \CTpr, \CFpr\}
  \]
  and the equality is attain iff $\CTpr = \CFpr$.
\end{customthm}

\proof
The value of $\Delta$ is defined as the maximal width of the interval $(\LB(\eta),\UB(\eta))$ w.r.t. $\eta$, that is,
\begin{equation}
 \label{equ:Delta}
  \Delta = \max_{\eta\in(0,1)}(\UB(\eta)-\LB(\eta))\:,
\end{equation}
where
\begin{eqnarray*}
\label{equ:lb}
 \displaystyle \LB(\eta) = \min_{\TPR\in\ITpr\atop{\FPR\in\IFpr}} \Prec(\eta,\TPR,\FPR)\,,\\
 \label{equ:ub}
\displaystyle \UB(\eta) = \max_{\TPR\in\ITpr\atop{\FPR\in\IFpr}} \Prec(\eta,\TPR,\FPR) \:,
\end{eqnarray*}
and
\begin{equation}
\label{equ:prec}
  \Prec(\eta,\TPR,\FPR) = \frac{\eta\cdot \TPR}{\eta\cdot\TPR+(1-\eta)\cdot \FPR}=\frac{1}{1+\frac{(1-\eta)}{\eta}\frac{\FPR}{\TPR}}\:.
\end{equation}
The equation~\eqref{equ:prec} shows that $\Prec(\eta,\TPR,\FPR)$ is a monotonically decreasing function of the ratio $\frac{\FPR}{\TPR}$ which together with $\FPR\in(\tFPR-\SigmaFpr,\tFPR+\SigmaFpr)$ and $\TPR\in(\tTPR-\SigmaTpr,\tTPR+\SigmaTpr)$ implies that
\[
   \UB(\eta) = \frac{1}{1+\frac{(1-\eta)}{\eta}\frac{\tFPR-\SigmaFpr}{\tTPR+\SigmaTpr}}\quad \mbox{and}\quad
   \LB(\eta) = \frac{1}{1+\frac{(1-\eta)}{\eta}\frac{\tFPR+\SigmaFpr}{\tTPR-\SigmaTpr}}\:.
\]
Using $x=\frac{1-\nu}{\nu}\in(0,\infty)$, we can re-parametrize $\UB(\eta)$ and $\LB(\eta)$ as
\[
   \UB(x) = \frac{1}{1+x\cdot r_1} \quad\mbox{and}   \quad 
   \LB(x) = \frac{1}{1+x\cdot r_2}
\]
where $r_1=\frac{\tFPR-\SigmaFpr}{\tTPR+\SigmaTpr}$ and $r_2=\frac{\tFPR+\SigmaFpr}{\tTPR-\SigmaTpr}$ \:. The value of $\Delta$ can be then equivalently defined as
\[
 \Delta = \max_{x\in(0,\infty)} f(x)\quad\mbox{and}\quad f(x)=\UB(x)-\LB(x)=\frac{1}{1+x\cdot r_1}-\frac{1}{1+x\cdot r_2} \:. 
\]
The maximum of $f(x)$ can be found analytically by solving $f'(x)=0$ for $x$ which yields
\[
   x^* = \frac{1}{\sqrt{r_1\cdot r_2}} \:,
\]
and hence 
\begin{equation}
  \label{equ:delta1}
   \Delta = f(x^*) = \frac{1}{1+\frac{r_1}{\sqrt{r_1\cdot r_2}}} - \frac{1}{1+\frac{r_2}{\sqrt{r_1\cdot r_2}}} 
   = \frac{1-\sqrt{\frac{r_1}{r_2}}}{1+\sqrt{\frac{r_1}{r_2}}}\:.
\end{equation}
It is seen that $\Delta$ is monotonically decreasing with the ratio
\begin{equation}
 \label{equ:cv1}
   \frac{r_1}{r_2} = \frac{\tFPR-\SigmaFpr}{\tTPR+\SigmaTpr}\cdot \frac{\tTPR-\SigmaTpr}{\tFPR+\SigmaFpr} = \frac{1-\CFpr}{1+\CFpr}\cdot\frac{1-\CTpr}{1+\CTpr} \:.
\end{equation}
If $\CTpr=\CFpr=\text{C}$ then 
\[
   \Delta=\frac{1-\sqrt{\frac{(1-C)^2}{(1+C)^2}}}{1+\sqrt{\frac{(1-C)^2}{(1+C)^2}}}
   = C = \max\{\CTpr,\CFpr\}\,.
\]
Decreasing either $\CTpr$ or $\CFpr$ increases the ratio $\frac{r_1}{r_2}$ resulting in the decrease of $\Delta$. Hence $\Delta\leq \max\{\CTpr,\CFpr\}$ and $\Delta=\max\{\CTpr,\CFpr\}$ iff $\CTpr=\CFpr$.

\eproof

\section{Exact relationship between $\Delta$ and coefficients of variation}

Assume the following practical problem. For a given sample of positive examples $\mathcal{X}^+$, we have computed $\tTPR$ along with the confidence interval width $\SigmaTpr$ and hence we could evaluate the corresponding coefficient of variation $\CTpr=\frac{\SigmaTpr}{\tTPR}$. We are about to collect sufficient amount of the negative examples $\mathcal{X}^-$ in order to estimate $\FPR$. What is the maximal coefficient of variation $\CFpr$ we can afford while still having the guarantee that the maximal uncertainty in measuring the precision is at most some prescribed $\Delta$ (for exact definition of $\Delta$ see~\eqref{equ:Delta}) ? Provided $\CTpr\leq \Delta$ we can use  Theorem~\ref{th:doundOnDelta1} to shows that the guarantee is met if $\CFpr=\Delta$. However, this estimate of $\CFpr$ is i) applicable only if $\CTpr\leq \Delta$ and ii) the value of $\CFpr$ can be unnecessary pessimistic if $\CTpr$ and $\CFpr$ are too different. 

\begin{figure}
    \centering
    \includegraphics[width=0.8\textwidth]{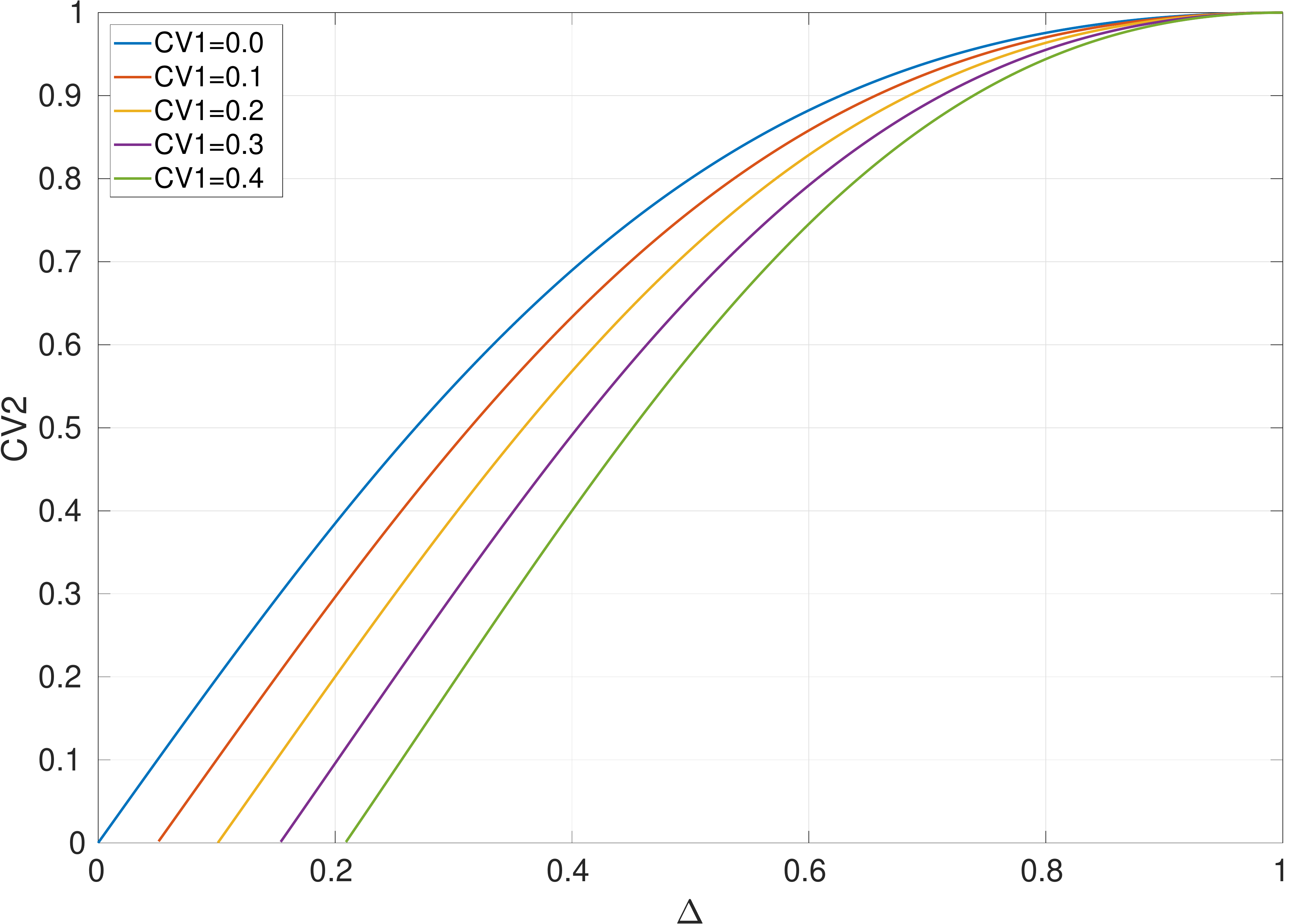}
    \caption{Dependence of one coefficient of variation $\C2\in\{\CTpr,\CFpr\}$ on the maximal uncertainty $\Delta$ while the other coefficient of variation $\C1\in\{\CTpr,\CFpr\}\setminus\{\C2\}$ is fixed.}
    \label{fig:cv1_cv2_delta}
\end{figure}

Thanks to~\eqref{equ:delta1} and~\eqref{equ:cv1}, the values of $\CFpr$, $\CTpr$ and $\Delta$ are related by equation
\[
\frac{1-\CFpr}{1+\CFpr}\cdot\frac{1-\CTpr}{1+\CTpr} = \left(  \frac{1-\Delta}{1+\Delta} \right )^2 \:.
\]
Note that the equation is symmetric with respect to $\CTpr$ and $\CFpr$. Hence for fixed $\Delta$ and $\C2\in\{\CTpr,\CFpr\}$ we can compute the value of $\C1\in\{\CTpr,\CFpr\}\setminus \{\C2\}$ exactly by
\[
   \C2 = \frac{(\C1+1)(1+k)-2}{(\C1+1)(1-k)-2}
   \quad\mbox{where}\quad k = \left( \frac{1-\Delta}{1+\Delta}\right )^2 \:.
\]
Figure~\ref{fig:cv1_cv2_delta} shows dependence of $\C2$ on the value of $\Delta$ when $\C1$ is fixed.

\end{subappendices}

\end{document}